\documentclass[11pt, a4paper]{article}

\usepackage[utf8]{inputenc}
\usepackage{amsmath}
\usepackage{amssymb} 
\usepackage{graphicx}
\usepackage{booktabs}
\usepackage{authblk}
\usepackage{geometry}
\usepackage{hyperref}
\usepackage{multicol}
\usepackage{caption} 
\title{Generic Vision and Cross-Attention for Reaction Yield Prediction}

\author[1,2]{Qiwei Han}
\author[2]{Chi Zhou}
\affil[1]{\small Department of Chemistry, Duke University, Durham, NC 27708, USA}
\affil[2]{\small Department of Computer Science, Georgia Institute of Technology, Atlanta, GA 30332, USA}
\date{}

\begin{document}
	
	\maketitle
	
	\begin{abstract}
		Traditional reaction yield prediction is constrained by 1D quantum descriptors that lack explicit spatial information. To address this gap, a dual-modal Vision Cross-Attention architecture is proposed, fusing tabular physical-organic data with 2D molecular topologies. Notably, it is demonstrated that a generic computer vision backbone processing simple 2D skeletal structures independently outperforms purely quantum-based baselines. By synergizing both modalities, superior predictive accuracy compared to traditional methodologies is achieved by the optimal cross-attention framework (Test RMSE = 5.27\%). Through mechanistic probing, active, descriptor-guided spatial querying is observed, effectively offloading macroscopic steric identification to the visual pathway. Furthermore, a dynamic chemical hierarchy is learned by the network to heavily prioritize critical steric bottlenecks, such as the aryl halide. Concurrently, residual skip connections are utilized to protect non-spatial electronic parameters from destructive attenuation during fusion. Collectively, a scalable and highly interpretable blueprint is provided for augmenting physical chemistry with deep visual learning.
		\vspace{2em}
	\end{abstract}
	
	\vspace{0.5cm}
	
	\begin{multicols}{2}
		
	\section{Introduction}
	
	The integration of high-throughput experimentation (HTE) with machine learning (ML) has fundamentally transformed the landscape of predictive synthetic chemistry \cite{ahneman2018predicting, perera2018platform, strieth2020machine, coley2020autonomous}. The ability to accurately predict reaction yields \textit{in silico} offers a paradigm shift for reaction optimization, enabling researchers to virtually screen vast combinatorial chemical spaces while drastically reducing the time, cost, and material waste associated with empirical bench-work \cite{shields2021bayesian, schwallko2021machine}. 
	
	Historically, state-of-the-art yield prediction models have relied heavily on explicit 1D molecular representations, such as Morgan fingerprints, reaction-driven fingerprints (DRFP), or explicitly calculated quantum-mechanical (QM) descriptors \cite{probst2022reaction, zahrt2019prediction, reid2019holistic, sandfort2020structure}. Seminal work by Doyle, Sigman, and others has demonstrated that concatenating molecular features—such as HOMO/LUMO energies, dipole moments, and Sterimol parameters—into tabular datasets can yield highly accurate predictive models when paired with algorithms like Random Forests or Multilayer Perceptrons (MLPs) \cite{ahneman2018predicting, gensch2022comprehensive}. However, this traditional tabular approach presents two significant bottlenecks. First, the explicit calculation of high-level Density Functional Theory (DFT) descriptors is computationally prohibitive for millions of theoretical screening candidates \cite{grambow2020deep}. Second, compressed 1D numerical arrays often struggle to holistically capture the complex, 2D and 3D steric environments and topological overlaps that govern transition-state reactivity in a reaction mixture \cite{kearnes2016molecular, wu2018moleculenet}.
	
	To bypass the computational expense of QM descriptors, recent advances in chemical deep learning have increasingly turned to implicit spatial representations, most notably through computer vision \cite{tetko2020bigchem, rajan2021decimer}. Inspired by the success of Convolutional Neural Networks (CNNs) in image recognition, researchers have demonstrated that networks operating directly on 2D skeletal molecular drawings (e.g., Chemception) can successfully extract chemical heuristics—such as aromaticity, steric hindrance, and functional group presence—without requiring prior physical calculations \cite{goh2017chemception, kimber2021deep, jimenez2020drug}. While these vision-only approaches offer extraordinary computational scalability, they inherently lack the explicit quantum-physical precision (e.g., exact atomic charges) provided by tabular data, often resulting in an informational ceiling when predicting highly sensitive catalytic yields \cite{janet2019quantitative, chuang2018comment}.
	
	Consequently, the next frontier in chemical AI lies in multimodal architectures capable of bridging the gap between implicit spatial topologies (vision) and explicit physical parameters (tabular data) \cite{guo2023multimodal}. To maximize this multimodal synergy, previous research has extensively explored cross-attention mechanisms, demonstrating them to be a highly promising direction for fusing disparate data types \cite{vaswani2017attention, jaegle2021perceiver, han2026chemfusion}. By allowing one modality to dynamically query the localized features of another, these attention-based architectures have consistently achieved state-of-the-art performance in complex multimodal tasks, making them the theoretical ideal for aligning 1D physical properties with 2D chemical structures. 
	
	In this work, we introduce a dual-modal architecture that synergizes explicit tabular QM data with high-quality 2D molecular images to predict high-throughput reaction yields. Specifically, we sought to determine if an off-the-shelf, generic vision backbone (ResNet-18)—completely devoid of domain-specific chemical pre-training—could extract sufficient topological information to rival explicitly calculated physics descriptors. Furthermore, we conducted a rigorous architectural ablation study to investigate the underlying mechanisms of multimodal fusion, directly comparing dynamic cross-attention against explicit vector concatenation (Simple Concat). 
	
	Notably, our results demonstrate that a generic vision model processing high-quality 2D reaction topologies significantly outperforms a purely tabular MLP baseline anchored on computationally expensive quantum-mechanical descriptors (5.60\% vs. 6.79\% RMSE). By unifying these modalities through our optimal Vision Cross-Attention framework, we successfully bridge 2D topology and 1D physical parameters to achieve superior predictive accuracy compared to traditional methodologies (Test RMSE = 5.27\%). 
	
	Beyond top-level predictive accuracy, we conducted a mechanistic analysis to understand the internal logic of this multimodal approach. Our findings indicate that rather than passively pooling visual data, the cross-attention mechanism actively queries the 2D images based on the tabular descriptors, allowing it to efficiently map macroscopic steric environments. Furthermore, analyzing the attention weights reveals that the network establishes a learned chemical hierarchy. It exhibits highly targeted spatial routing, prioritizing critical steric bottlenecks (e.g., the aryl halide) while de-emphasizing the topology of non-critical components. Finally, we observed a "Residual Bottleneck," highlighting that skip connections are mathematically necessary to preserve purely non-spatial quantum parameters during the fusion process. Overall, this framework provides a highly interpretable, computationally efficient blueprint for augmenting explicit physical chemistry with generic computer vision.

\section{Computational Methods}

\begin{figure*}[t]
	\centering
	\includegraphics[width=0.9\textwidth]{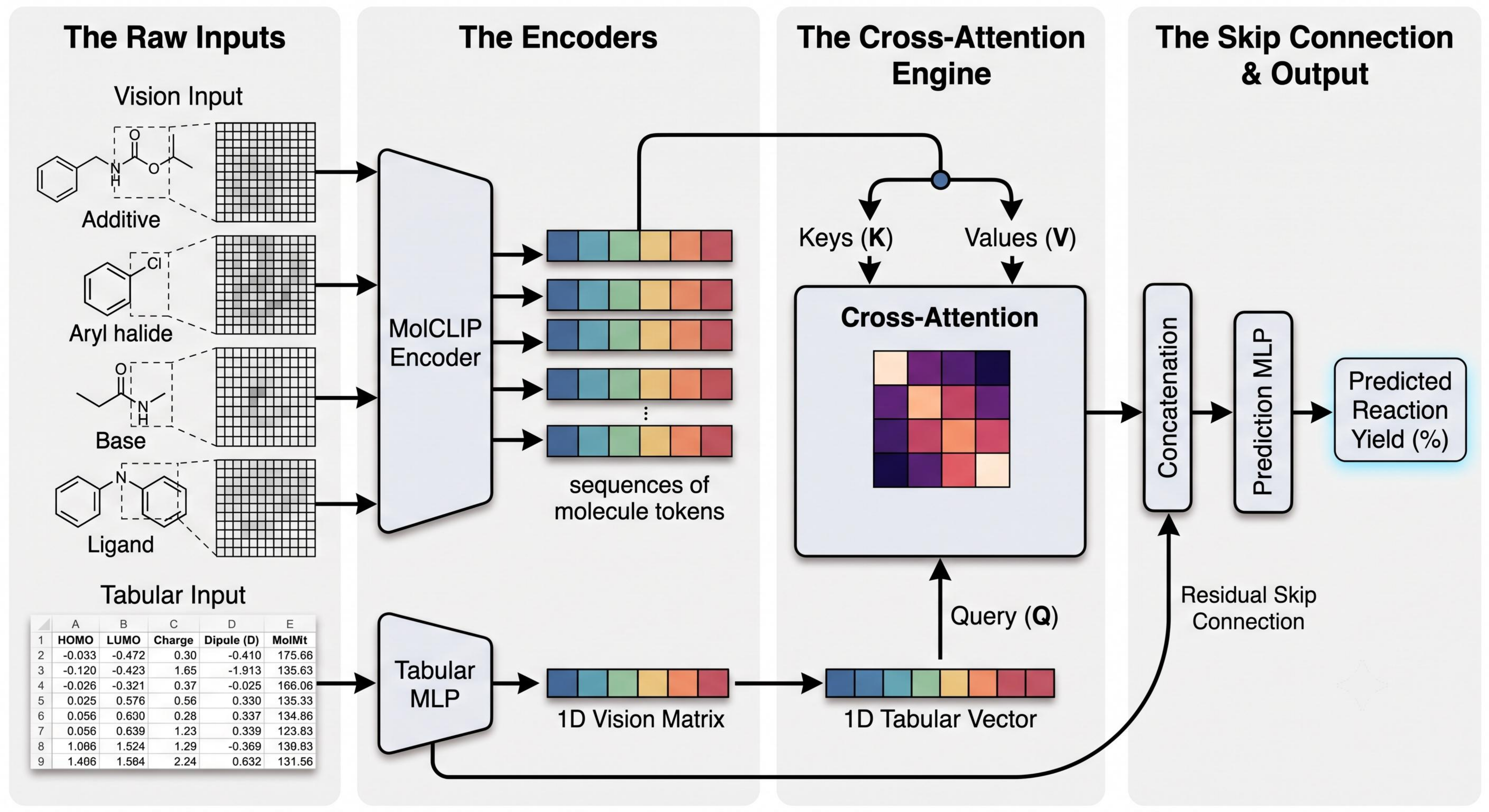}
	\caption{\textbf{The Dual-Modal Vision Cross-Attention Architecture.} 2D spatial topologies extracted via a generic ResNet-18 backbone are actively queried by 1D physical-organic descriptors to dynamically contextualize the macroscopic reaction space. A residual skip connection explicitly bypasses the spatial attention bottleneck to preserve strict quantum-chemical rigor prior to the final yield prediction.}
	\label{fig:architecture}
\end{figure*}

\subsection{Dataset Partitioning and Modality Generation}
We utilized the benchmark \textit{rxnpredict} dataset developed by the Doyle Lab \cite{ahneman2018predicting}, which maps high-throughput reaction yields to rigorously calculated tabular physical-organic descriptors (e.g., HOMO/LUMO energies, dipole moments, and Sterimol parameters). To ensure rigorous evaluation, the dataset was randomly partitioned into a 70:30 training and held-out test split, utilizing a locked random seed to guarantee strict reproducibility across all architectural ablations. 

To facilitate the dual-modal architecture, each reaction was parsed into two distinct representations:
\begin{itemize}
	\item \textbf{Tabular Descriptors:} The explicit 1D physical-organic features were extracted and normalized using zero-mean, unit-variance standardization fitted exclusively on the training set to prevent data leakage.
	\item \textbf{2D Visual Topologies:} We generated standard 2D skeletal structure images for each individual reaction component—specifically the additive, aryl halide, base, and ligand—using RDKit \cite{rdkit}. Images were standardized to a resolution of $224 \times 224$ pixels, converted to RGB tensors, and normalized using standard ImageNet parameters. To ensure the network learned innate chemical topology rather than the arbitrary layout of a composite grid, these components were rendered and processed as independent images, aligned during training via a unique compound key (experimental plate, row, and column).
\end{itemize}

\subsection{Modality Encoders}
To process the multi-modal inputs, we employed distinct neural network pathways for feature extraction prior to modality fusion.

\vspace{2mm}
\noindent \textbf{Tabular Pathway:} The scaled physical-organic descriptors were passed through a Feed-Forward Multi-Layer Perceptron (MLP) to generate a dense, dimensionally aligned tabular embedding of size $d=256$.

\vspace{2mm}
\noindent \textbf{Vision Pathway:} To extract 2D spatial features, we purposefully utilized a generic ResNet-18 convolutional neural network \cite{he2016deep} pre-trained exclusively on ImageNet. While domain-specific models are tailored for physical chemistry, this backbone was designed for general-purpose computer vision and was not specifically exposed to molecules during pre-training. We hypothesized that the network could implicitly learn 2D steric bulk and topological overlaps strictly from the geometric arrangement of the line-angle drawings. To isolate the cross-attention learning dynamics, the convolutional backbone parameters were completely frozen. The four reactant images were processed individually and stacked along the sequence dimension, yielding a discrete visual embedding matrix $X_{vis} \in \mathbb{R}^{4 \times d_{k}}$, allowing the downstream fusion mechanisms to interact with each specific reagent independently. 

\subsection{Architectural Ablation and Fusion Strategies}
To rigorously investigate the underlying mechanisms of feature fusion, we systematically evaluated two distinct mathematical approaches for bridging the explicit 1D tabular data with the implicit 2D visual embeddings: an explicit concatenation approach (Simple Concat) and a dynamic querying approach (Cross-Attention).

\vspace{2mm}
\noindent \textbf{Strategy A: Simple Concat Fusion (The Baseline)} \\
In the Simple Concat architecture, we bypass complex spatial routing in favor of a chemically blind, globally averaged topological state. The component visual embeddings ($X_{vis}$) are collapsed via an unweighted Mean Pooling operation:
\begin{equation}
	X_{pool} = \frac{1}{4} \sum_{j=1}^{4} X_{vis, j}
\end{equation}
This averaged visual state is then explicitly concatenated with the raw tabular descriptors ($X_{tab}$) and passed directly to the prediction head:
\begin{equation}
	X_{fused} = \text{Concat}(X_{pool}, X_{tab})
\end{equation}

\vspace{2mm}
\noindent \textbf{Strategy B: Multi-Head Cross-Attention Fusion (The Champion)} \\
To test whether dynamic, descriptor-guided spatial weighting could outperform a rigid global average, we designed a Multi-Head Vision Cross-Attention architecture (Figure \ref{fig:architecture}). In standard attention paradigms, relevance scores are computed via the dot product of a Query ($Q$) and a Key ($K$), scaled and multiplied by a Value ($V$) \cite{vaswani2017attention}. 

In our proposed architecture, the scaled physical-organic tabular descriptors ($X_{tab}$) act as the central anchor (the Query), while the 2D visual embeddings extracted from the generic ResNet-18 backbone ($X_{vis}$) act as the Keys and Values. This explicitly allows the quantum math to actively search the visual topology. To enable the network to simultaneously monitor distinct topological features, we execute spatial routing via Multi-Head Cross-Attention utilizing $h=8$ independent heads. For each head $i \in \{1, \dots, h\}$, we compute the projections into a shared latent dimension $d_k=256$ using learnable weight matrices $W_{Q,i}$, $W_{K,i}$, and $W_{V,i}$:

\begin{align}
	Q_i &= X_{tab} W_{Q,i} \\
	K_i &= X_{vis} W_{K,i} \\
	V_i &= X_{vis} W_{V,i}
\end{align}

The cross-attention for each independent head is subsequently computed by taking the scaled dot-product of the explicit tabular query and the implicit visual keys, generating dynamic spatial routing weights:
\begin{equation}
	\text{head}_i = \text{softmax}\left(\frac{Q_i K_i^T}{\sqrt{d_k}}\right)V_i
\end{equation}

The outputs from all $h=8$ heads are then concatenated and projected through a final output weight matrix $W^O$ to produce the unified vision-contextualized embedding, $X_{attn}$:
\begin{equation}
	X_{attn} = \text{Concat}(\text{head}_1, \dots, \text{head}_h)W^O
\end{equation}

Crucially, to prevent this spatial attention mechanism from acting as a destructive informational bottleneck for purely non-spatial parameters (e.g., electronic states or dipole moments), we implemented a residual skip-connection \cite{he2016deep}. The contextualized output from the multi-head attention block ($X_{attn}$) is concatenated directly with the raw, unmodified tabular features ($X_{tab}$) to shield the rigorous quantum chemistry:

\begin{equation}
	X_{fused} = \text{Concat}(X_{attn}, X_{tab})
\end{equation}

Finally, for both architectural strategies, this fused representation is passed through a Multi-Layer Perceptron (MLP) prediction head to output the continuous reaction yield prediction, $\hat{y}$:
\begin{equation}
	\hat{y} = \text{MLP}(X_{fused})
\end{equation}

\subsection{Training Protocol and Interpretability Metrics}
All networks were implemented in PyTorch and trained to predict continuous reaction yield (\%) utilizing a Mean Squared Error (MSE) loss function. Optimization was driven by the AdamW optimizer (batch size = 32, initial learning rate = $1 \times 10^{-3}$, weight decay = $1 \times 10^{-4}$), with the learning rate dynamically modulated via a Cosine Annealing scheduler with warm restarts. Models were trained for 500 epochs, employing strict checkpointing to preserve the specific network weights that minimized held-out validation loss.

Top-level predictive performance was quantified on the unseen test set using Root Mean Square Error (RMSE) and the coefficient of determination ($R^2$). Beyond predictive accuracy, we extracted specific interpretability metrics to determine mechanistic feature reliance. Permutation Feature Importance (PFI) was calculated by individually shuffling each tabular feature array across the test set and recording the absolute resulting increase in RMSE. Finally, for the spatial routing analysis, raw softmax attention scores were extracted directly from the cross-attention layers during the forward pass over the test set, prior to multiplication with the Value ($V$) matrix.
		
	\section{Results and Discussion}
	
	\subsection{The Surprising Efficacy of Generic Vision and Modality Fusion}
	
	\begin{figure*}[htbp]
		\centering
		\includegraphics[width=1.0\linewidth]{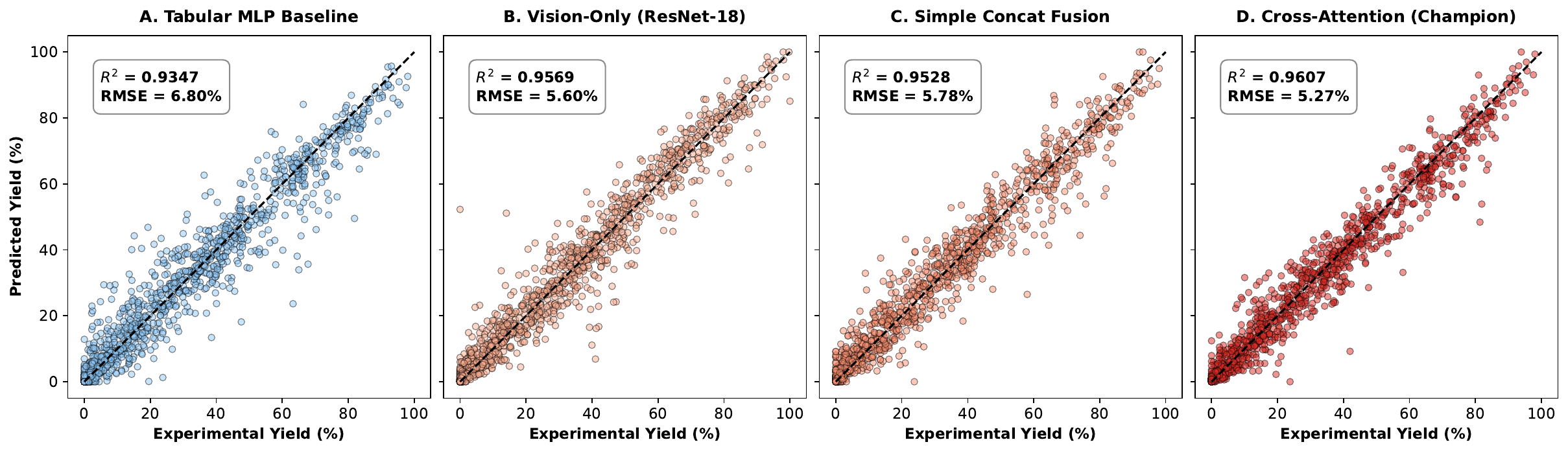} 
		\caption{\textbf{Evaluating Architectural Fusion and Modality Synergy.} Predictive performance (RMSE and $R^2$) across baseline and fused architectures. The dual-modal Cross-Attention framework achieves the lowest error, seamlessly synergizing spatial visual topology with explicit tabular quantum descriptors.}
		\label{fig:fusion_performance}
	\end{figure*}
	
	We first sought to answer a provocative core hypothesis: can an off-the-shelf, generic vision model extract enough spatial topology from high-quality 2D skeletal drawings to rival explicitly calculated quantum-mechanical descriptors? 
	
	As shown in Figure \ref{fig:fusion_performance}, the isolated Tabular MLP establishes a formidable baseline (RMSE = 6.80\%) by leveraging explicit 1D physical-organic features (e.g., HOMO/LUMO, dipole moments). Notably, this neural network baseline alone already significantly outperforms the previous state-of-the-art Random Forest benchmark established in the literature \cite{ahneman2018predicting}. However, when we evaluated the isolated Vision-Only baseline—where high-resolution 2D reaction topologies were processed through a frozen, ImageNet-trained ResNet-18 and aggregated via global mean pooling prior to final MLP prediction—the model achieved a significantly lower error (RMSE = 5.60\%). 
	
	This is a noteworthy observation. It implies that in complex chemical reactions, macroscopic steric environments are complementary and equally vital to electronic states, though they often remain difficult to capture using purely 1D quantum arrays. Conversely, the generic ResNet-18 model appears adept at extracting geometric bulk directly from the images, despite lacking explicit chemical pre-training.
	
	Interestingly, when we attempted a naive multimodal fusion (the Simple Concat architecture), the predictive performance was suboptimal, proving slightly worse than the Vision-Only baseline. This degradation highlights a well-known pitfall in multimodal learning: modality dominance \cite{wang2020what, wu2022characterizing}. When passively pooled visual embeddings are directly concatenated with dense, explicit tabular parameters, the network naturally gravitates toward the easily interpretable tabular numbers during training. Consequently, it dilutes the highly effective spatial representations learned by the vision backbone. 
	
	To overcome this modality dominance, our Cross-Attention framework eschews passive pooling in favor of an active, dynamic integration strategy \cite{vaswani2017attention, jaegle2021perceiver}. Rather than treating the 1D quantum arrays and 2D visual embeddings as isolated vectors to be concatenated, the cross-attention mechanism projects them into a shared latent space. Specifically, the explicit tabular descriptors act as targeted mathematical \textit{queries}, while the dense 2D spatial feature maps serve as the \textit{keys} and \textit{values}. This architectural bottleneck forces a strict inter-modal dependency: the network cannot simply ignore the visual data during training because the 1D electronic features are explicitly tasked with retrieving complementary steric information from the 2D image pathway. By structurally enforcing the equal importance of sterics and electronics without allowing one to overshadow the other, the framework successfully synergizes 1D physical properties with 2D spatial context, achieving a highly competitive predictive accuracy (5.27\% RMSE).
	
	\subsection{Mechanisms of Fusion: Shifting from Mass Proxies to Explicit Spatial Querying}
	\label{sec:mechanisms}
	
	\begin{figure*}[htbp]
		\centering
		\includegraphics[width=1.0\linewidth]{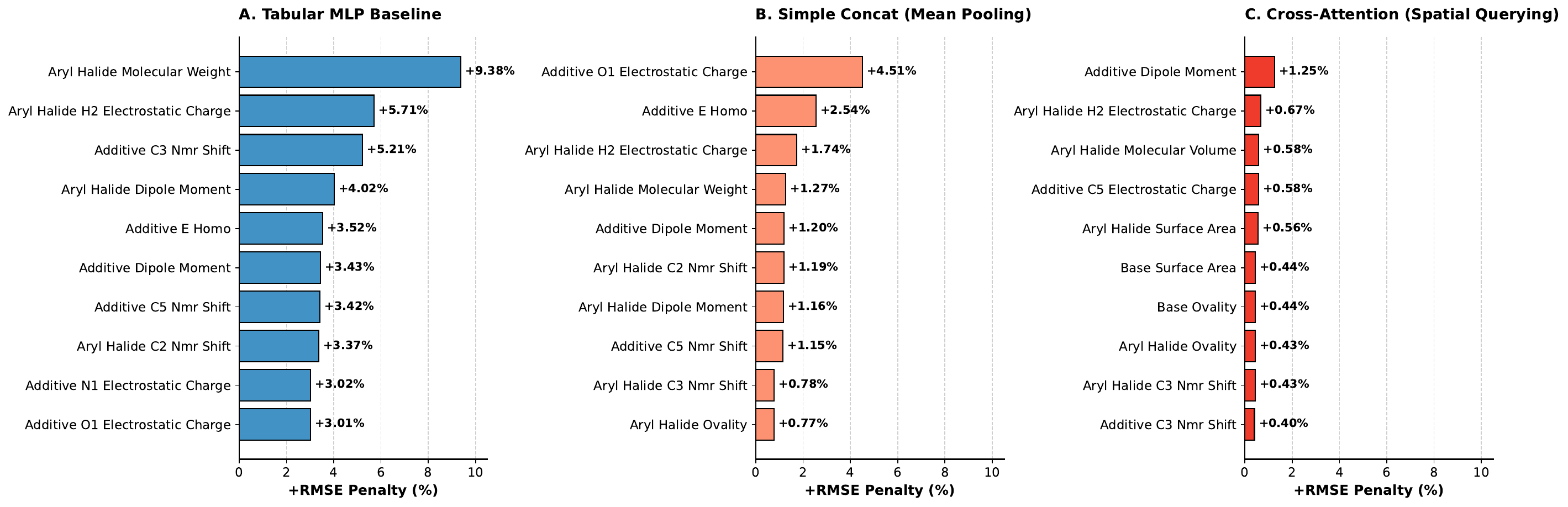}
		\caption{\textbf{Mechanistic shifts in feature reliance via Permutation Feature Importance (PFI).} \textbf{(A) Tabular Baseline:} The MLP relies heavily on \textit{aryl halide molecular weight} as a mathematical proxy for molecular size. \textbf{(B) Simple Concat:} Passive mean pooling blurs spatial details, forcing the network to take a computational shortcut, falling back on molecular weight and purely electronic descriptors. \textbf{(C) Cross-Attention:} The mechanism executes a targeted spatial alignment. By actively anchoring on explicit tabular shape parameters (e.g., volume, ovality, surface area), it successfully offloads steric identification to the 2D visual embeddings.}
		\label{fig:pfi_mechanisms}
	\end{figure*}
	
	To elucidate the internal reasoning of the networks beyond top-level performance metrics, we employed Permutation Feature Importance (PFI) \cite{fisher2019all} to map how the different architectures allocate predictive labor between the explicit quantum chemistry and the computer vision pathway (Figure \ref{fig:pfi_mechanisms}).
	
	In the isolated Tabular MLP (Figure \ref{fig:pfi_mechanisms}A), the network exhibits a disproportionate reliance on \texttt{aryl\_halide\_molecular\_weight} (+9.38\% RMSE penalty). Because the strictly 1D architecture lacks macroscopic spatial context, it is forced to utilize molecular weight as a general, mathematical heuristic to estimate steric bulk. 
	
	Previously, we attributed the performance degradation of the Simple Concat architecture relative to the Vision-Only baseline to modality dominance. By analyzing its Permutation Feature Importance (Figure \ref{fig:pfi_mechanisms}B), we uncover the precise computational behavior driving this dominance: \textit{shortcut learning}. Because the simple concatenated framework compresses the ResNet feature maps using unweighted mean pooling, critical localized spatial details are mathematically obscured. Unlike the isolated Vision-Only baseline, which is strictly forced to decipher visual features to minimize loss, the concatenated model is provided an alternative: the explicit tabular descriptors. Consequently, the network takes a computational shortcut, preferentially extracting its predictive signal from the mathematically simpler tabular features rather than untangling the blurred visual embeddings. This dynamic explains the shifting, yet persistent, importance of the aryl halide molecular weight. Although the introduction of macroscopic visual data reduces its absolute dominance—dropping it from the top-ranked feature to the fourth—molecular weight remains firmly within the top ten. It survives as a necessary 1D heuristic proxy for steric bulk precisely because the network's visual spatial resolution has been degraded, illustrating why simple concatenation struggles to fully synergize multimodal data.
	
	In stark contrast, the Cross-Attention framework (Figure \ref{fig:pfi_mechanisms}C) executes a highly active, \textit{descriptor-guided spatial alignment}. Strikingly, molecular weight completely vanishes from its most critical features. Instead, the tabular query relies strictly on explicit macroscopic shape descriptors—specifically \texttt{molecular\_volume}, \texttt{surface\_area}, and \texttt{ovality}. The cross-attention mechanism actively utilizes these physical parameters to query the generic ResNet topology, effectively tasking the visual pathway to localize the corresponding steric density within the 2D image. This demonstrates that the cross-attention architecture successfully and selectively offloads steric identification to the visual modality, rendering the fusion model exceptionally robust to tabular permutations.
	
	A critical question arises: why does the isolated Tabular MLP rely on molecular weight rather than the explicit geometric descriptors (e.g., volume) already present in the tabular dataset? This behavior stems from the interplay of feature variance and spatial blindness. Molecular weight is an exact, invariant scalar, whereas computed molecular volume is intrinsically noisy, heavily dependent on the specific 3D conformer generation protocol \cite{ebejer2012freely, axelrod2022geom}. Because a purely 1D architecture must treat its inputs as definitive answers—and lacks the spatial context to localize where an estimated volume physically resides—it preferentially anchors to the noiseless exactitude of molecular mass to approximate generalized bulk. 
	
	However, within the Cross-Attention framework, the role of tabular volume fundamentally shifts: it is no longer treated as a definitive scalar answer, but rather as a directional search query. This aligns perfectly with the formal Query-Key-Value (QKV) routing established in modern attention mechanisms \cite{vaswani2017attention, jaegle2021perceiver, carion2020end}. Because geometric parameters like volume and surface area correlate directly with 2D topological area, the network utilizes these noisy tabular estimates as explicit mathematical queries to actively interrogate the visual pathway. The high-fidelity 2D image serves as the absolute structural verifier. The model uses the rough volume estimate to search the visual keys, relying entirely on the image's values to confirm, localize, and map the actual steric bulk. By utilizing the visual modality to ground the noisy geometric query in absolute spatial reality, the cross-attention mechanism successfully mitigates tabular conformer noise and renders the crude proxy of molecular weight obsolete.
	
	Finally, it is worth considering the behavior of the cross-attention mechanism in data-constrained scenarios where explicit geometric tabular descriptors (e.g., volume and surface area) are omitted from the dataset. In such cases, neural networks exhibit opportunistic feature routing. Deprived of direct geometric queries, the tabular encoder synthesizes a latent spatial query by aggregating secondary 1D proxies, such as molecular weight, heavy atom count, and implicit electronic signatures. While this synthesized query is inherently less precise than explicit volume, the architectural integrity of the cross-attention bottleneck remains intact. The network still projects this proxy-based query into the shared latent space to interrogate the visual pathway, allowing the 2D image to continuously function as the absolute structural verifier—compensating for the impoverished tabular query by mapping crude mass approximations directly onto the ground-truth visual topology.
		
	\subsection{Architectural Ablation and the Learned Chemical Hierarchy}
	\label{sec:unpacking_champion}
	
	\begin{figure*}[htbp]
		\centering
		\includegraphics[width=1.0\linewidth]{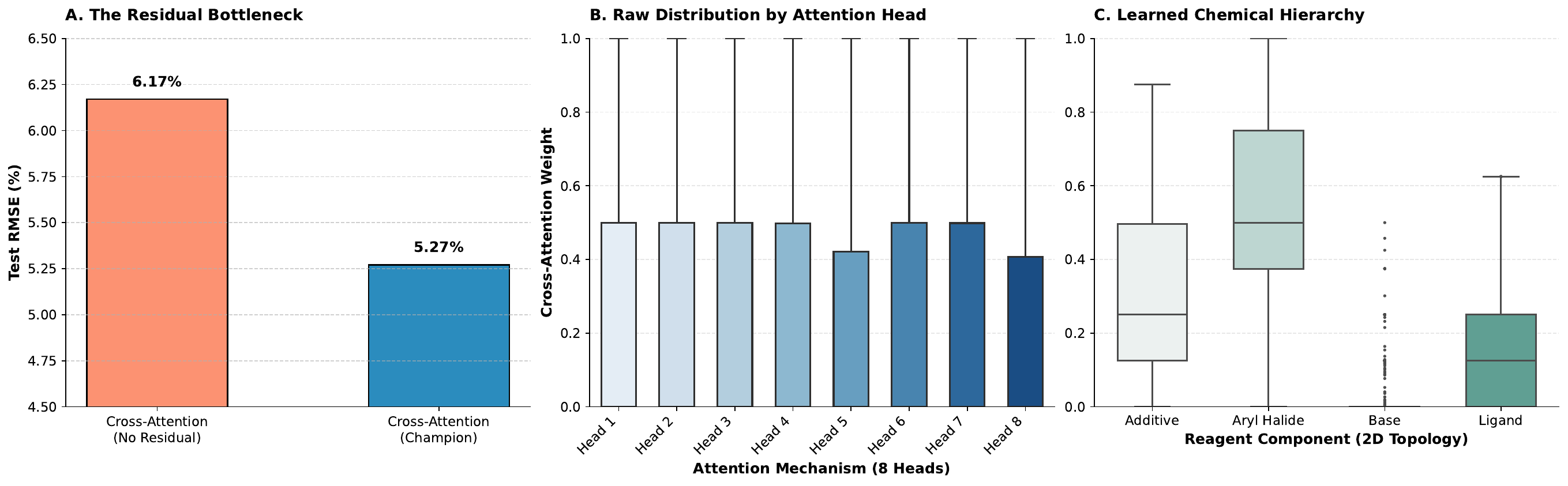}
		\caption{\textbf{Architectural ablation and visual verification of dynamic spatial specialization.} \textbf{(A) The Residual Bottleneck:} Forcing non-spatial electronic descriptors entirely through the spatial attention block severely compromises predictive accuracy (6.17\% RMSE). Implementing a residual skip connection protects these quantum parameters from destructive attenuation, restoring superior predictive accuracy (5.27\%). \textbf{(B) Raw Distribution by Head:} The raw softmax attention scores from the 8 independent heads exhibit extreme variance (0.0 to 1.0), demonstrating sparse, binary-like spatial routing rather than passive, uniform pooling. \textbf{(C) Learned Chemical Hierarchy:} Tokenizing the visual input into four distinct reactant embeddings enables a direct mapping of attention scores to specific chemical components. The network dynamically shifts its visual focus, heavily prioritizing the 2D topology of the Aryl Halide (Mean = 0.5327) while almost entirely suppressing the Base (Mean = 0.0294).}
		\label{fig:attention_specialization}
	\end{figure*}
	
	To understand the specific design principles contributing to our champion 5.27\% RMSE, we performed a targeted architectural ablation, investigating both the information flow and the spatial attention distribution (Figure \ref{fig:attention_specialization}).
	
	We first identified a critical phenomenon termed the \textit{Residual Bottleneck} (Figure \ref{fig:attention_specialization}A). When explicit quantum descriptors are exclusively forced through the attention fusion block (Cross-Attention without Residual), the model exhibits a severe accuracy penalty (RMSE = 6.17\%). This proves that while attention acts as an effective filter for querying physical shapes, it acts as a destructive informational bottleneck for purely non-spatial quantum arrays (e.g., dipole moments or HOMO energies). The inclusion of a residual skip connection mathematically bypasses this bottleneck, shielding the unattenuated chemistry and culminating in the highly linear predictive correlation of the Champion architecture.
	
	Finally, to demystify how the network routes this spatial information, we systematically extracted the model's internal attention mechanisms during the evaluation phase. Specifically, we performed a complete forward pass over the unseen held-out test set. To ensure we captured the network's pure spatial decision-making process, we extracted the data using the standard scaled dot-product attention formulation:
	$$Attention\ Weights = Softmax\left(\frac{Q \times K^T}{\sqrt{d}}\right)$$
	By logging the raw attention tensors strictly at this computational step—immediately following the softmax normalization but prior to multiplication with the visual Value ($V$) matrix—we isolate the unadulterated probability distributions generated by the 8 independent heads. Unlike static model parameters, these extracted scores sum to 1.0 and dictate exactly where the tabular query ($Q$) mathematically "looks" among the visual keys ($K$) for any given reaction. 
	
	The global distribution of these raw attention tensors across the test set (Figure \ref{fig:attention_specialization}B) reveals extreme variance. Rather than defaulting to a uniform baseline, individual heads frequently output scores of absolute zero (Median = 0.0000), indicating that the cross-attention mechanism executes highly sparse, binary-like spatial routing.
	
	Furthermore, because the visual pathway processes the reaction as four distinct image embeddings loaded in strict sequence (Additive, Aryl Halide, Base, and Ligand), we can directly map these logged attention probabilities back to specific chemical components. By averaging the raw attention distributions across all 8 independent heads for every reaction in the test set, we calculated a singular global attention map for each prediction. This aggregated data allowed us to quantify the network's overall visual priority, establishing the learned chemical hierarchy shown in Figure \ref{fig:attention_specialization}C.
	
	Statistical analysis of these component-mapped distributions refutes the hypothesis that the generic ResNet embeddings lack the resolution for targeted spatial querying. The network entirely avoids passive mean pooling—which would yield a uniform 0.25 attention score with zero variance across all components. Instead, it establishes a strict, data-driven chemical hierarchy. The network heavily prioritizes the 2D topology of the Aryl Halide, allocating it 53.27\% of the global attention budget on average (Mean = 0.5327). Strikingly, the Aryl Halide's median attention score is 0.5000, indicating that for at least half of all test reactions, the network dedicates the absolute majority of its visual focus exclusively to this single structural bottleneck. Conversely, the 2D topology of the Base is completely suppressed (Mean = 0.0294), with the network allocating it zero attention in over 75\% of the dataset (75th Percentile = 0.0000). 
	
	This highly polarized, sparse distribution proves that the cross-attention mechanism is profoundly dynamic. Guided by the explicit tabular quantum descriptors, the AI effectively learned which specific steric bottlenecks govern the reaction space, actively routing its generic visual pathway to the most critical structural components.
		
	\section{Limitations and Future Directions}While the dual-modal Cross-Attention architecture achieves highly competitive predictive performance, several limitations present clear avenues for future research. First, although the generic ResNet-18 excels at macroscopic spatial routing, it lacks explicit atomic precision. Substituting this backbone with chemically pre-trained vision foundation models could enable the microscopic localization of specific functional groups or transition-state geometries. Second, the framework relies on computationally expensive DFT-derived quantum descriptors (e.g., HOMO/LUMO energies) to anchor its queries, limiting the high-throughput screening of uncharted molecular scaffolds. Future iterations could integrate Graph Neural Networks (GNNs) to rapidly approximate these quantum parameters, bypassing the DFT computational bottleneck. Finally, while 2D topologies effectively capture general steric bulk, they inherently lack dynamic 3D conformational awareness. Adapting this multimodal framework to incorporate computationally lightweight 3D point clouds or E(3)-equivariant architectures remains a critical next step for modeling highly stereoselective reaction pathways.
	
	\section{Conclusion}
	
	In this study, we developed a dual-modal Vision Cross-Attention architecture that successfully bridges the gap between rigorous, explicitly calculated quantum-chemical parameters and implicit 2D molecular topologies for high-throughput reaction yield prediction. By utilizing a parallel cross-attention mechanism coupled with a protective residual skip connection, our framework effectively fuses 1D physical-organic tabular descriptors with 2D spatial embeddings extracted via an off-the-shelf computer vision backbone (ResNet-18).
	
	Comprehensive evaluation on the benchmark \textit{rxnpredict} dataset yielded a compelling discovery: despite possessing strictly zero chemical pre-training, a generic vision model processing high-fidelity 2D geometric topologies significantly outperformed a standard baseline relying exclusively on computationally expensive quantum-mechanical descriptors (5.60\% vs. 6.79\% Test RMSE). Ultimately, unifying these representations in our optimal Cross-Attention architecture achieved superior predictive accuracy compared to traditional methodologies (RMSE = 5.27\%, $R^2$ = 0.961), systematically resolving complex edge cases that isolated, single-modality baselines struggled to contextualize.
	
	Beyond top-level performance metrics, our mechanistic analyses illuminated the internal logic of multimodal chemical networks. Through Permutation Feature Importance (PFI), we observed an elegant division of labor: the cross-attention mechanism actively anchors on tabular shape descriptors (e.g., volume and ovality) to query the visual embeddings, effectively offloading macroscopic steric identification to the 2D pathway. This frees the tabular network to focus exclusively on highly precise, non-spatial electronic states.
	
	Furthermore, extracting the raw softmax attention scores debunked the assumption that generic vision models force chemical networks into passive pooling. Instead, guided by explicit tabular queries, the network executes extreme, sparse spatial routing to establish a strict chemical hierarchy. It dynamically dedicated the vast majority of its visual focus to critical steric bottlenecks (e.g., the Aryl Halide) while completely suppressing the visual topology of non-critical components. Finally, our architectural ablations demonstrated the absolute necessity of residual skip connections to bypass this spatial attention block, preventing the destructive attenuation of non-spatial quantum parameters—a phenomenon we termed the \textit{Residual Bottleneck}.
	
	Ultimately, this work challenges the prevailing assumption that highly specialized, domain-specific pre-training is an absolute prerequisite for chemical deep learning. By demonstrating that generic visual networks can execute highly intelligent spatial routing when guided by explicit physical-organic mathematics, we provide a synergistic, computationally lightweight blueprint for the ongoing development of fast, spatially aware predictive models in synthetic reaction optimization.
		
		\section*{Acknowledgements}
		
		This research received no specific grant from any funding agency in the public, commercial, or not-for-profit sectors. The authors gratefully acknowledge the Doyle Lab at Princeton University for generating and open-sourcing the \textit{rxnpredict} dataset, which served as the foundational benchmark for this study. 
		
		\subsection*{Artificial Intelligence Disclosure}
		In accordance with academic publishing guidelines, the authors disclose the use of Google's Gemini as an editorial assistant for text refinement and to assist in generating the schematic in Figure 1. The authors take full responsibility for the final scientific content. All quantitative data visualizations (Figures 2, 3, and 4) were generated programmatically via Python directly from raw experimental and model outputs.
		\bibliographystyle{unsrt}
		\bibliography{references} 

@article{ahneman2018predicting, author={Ahneman, Derek T and others}, title={Predicting reaction performance in C--N cross-coupling using machine learning}, journal={Science}, volume={360}, pages={186--190}, year={2018}}

@article{perera2018platform, author={Perera, Damith and others}, title={A platform for automated nanomole-scale reaction screening and micromole-scale synthesis in flow}, journal={Science}, volume={359}, pages={429--434}, year={2018}}

@article{strieth2020machine, author={Strieth-Kalthoff, Felix and others}, title={Machine learning the ropes: principles, applications and directions in synthetic chemistry}, journal={Chem. Soc. Rev.}, volume={49}, pages={6154--6168}, year={2020}}

@article{coley2020autonomous, author={Coley, Connor W and others}, title={Autonomous discovery in the chemical sciences part II: Outlook}, journal={Angew. Chem. Int. Ed.}, volume={59}, pages={23414--23436}, year={2020}}

@article{shields2021bayesian, author={Shields, Benjamin J and others}, title={Bayesian reaction optimization as a tool for chemical synthesis}, journal={Nature}, volume={590}, pages={89--96}, year={2021}}

@article{schwallko2021machine, author={Schwallko, J and others}, title={Machine learning for chemical reactivity and reaction conditions}, journal={Curr. Opin. Chem. Biol.}, volume={65}, pages={11--18}, year={2021}}

@article{probst2022reaction, author={Probst, Daniel and others}, title={Reaction classification and yield prediction using the differential reaction fingerprint DRFP}, journal={Digit. Discov.}, volume={1}, pages={91--97}, year={2022}}

@article{zahrt2019prediction, author={Zahrt, Andrew F and others}, title={Prediction of higher-selectivity catalysts by computer-driven workflow and machine learning}, journal={Science}, volume={363}, pages={eaau5631}, year={2019}}

@article{reid2019holistic, author={Reid, Jolene P and others}, title={Holistic predictive models of spatial and electronic requirements for ligand-mediated selective catalysis}, journal={Nature}, volume={571}, pages={343--348}, year={2019}}

@article{sandfort2020structure, author={Sandfort, Frederik and others}, title={A structure-based platform for predicting chemical reactivity}, journal={Chem}, volume={6}, pages={1379--1390}, year={2020}}

@article{gensch2022comprehensive, author={Gensch, Tobias and others}, title={A comprehensive discovery platform for organophosphorus ligands for catalysis}, journal={J. Am. Chem. Soc.}, volume={144}, pages={1205--1217}, year={2022}}

@article{grambow2020deep, author={Grambow, Colin A and others}, title={Deep learning of activation energies}, journal={J. Phys. Chem. Lett.}, volume={11}, pages={2992--2997}, year={2020}}

@article{kearnes2016molecular, author={Kearnes, Steven and others}, title={Molecular graph convolutions: moving beyond fingerprints}, journal={J. Comput.-Aided Mol. Des.}, volume={30}, pages={595--608}, year={2016}}

@article{wu2018moleculenet, author={Wu, Zhenqin and others}, title={MoleculeNet: a benchmark for molecular machine learning}, journal={Chem. Sci.}, volume={9}, pages={513--530}, year={2018}}

@article{tetko2020bigchem, author={Tetko, Igor V and others}, title={BIGCHEM: Challenges and opportunities for big data analysis in chemistry}, journal={Mol. Inform.}, volume={39}, pages={1900132}, year={2020}}

@article{rajan2021decimer, author={Rajan, Kohulan and others}, title={DECIMER: towards deep learning for chemical image recognition}, journal={J. Cheminf.}, volume={13}, pages={1--9}, year={2021}}

@article{goh2017chemception, author={Goh, Garrett B and others}, title={Chemception: a deep neural network with minimal chemistry knowledge}, journal={arXiv preprint arXiv:1706.06689}, year={2017}}

@article{kimber2021deep, author={Kimber, Thomas B and others}, title={Deep learning in virtual screening: recent applications and developments}, journal={Int. J. Mol. Sci.}, volume={22}, pages={4435}, year={2021}}

@article{jimenez2020drug, author={Jim{\'e}nez-Luna, Jos{\'e} and others}, title={Drug discovery with explainable artificial intelligence}, journal={Nat. Mach. Intell.}, volume={2}, pages={573--584}, year={2020}}

@article{janet2019quantitative, author={Janet, Jon Paul and others}, title={A quantitative uncertainty metric controls error in neural network-driven chemical discovery}, journal={Chem. Sci.}, volume={10}, pages={7913--7922}, year={2019}}

@article{chuang2018comment, author={Chuang, Keiser V and others}, title={Comment on "Predicting reaction performance in C--N cross-coupling using machine learning"}, journal={Science}, volume={362}, pages={eaat8603}, year={2018}}

@article{guo2023multimodal, author={Guo, Jian and others}, title={Multimodal deep learning for chemical applications}, journal={J. Chem. Inf. Model.}, volume={63}, pages={3561--3578}, year={2023}}

@article{vaswani2017attention, author={Vaswani, Ashish and others}, title={Attention is all you need}, journal={Adv. Neural Inf. Process. Syst.}, volume={30}, year={2017}}

@inproceedings{jaegle2021perceiver, author={Jaegle, Andrew and others}, title={Perceiver: General perception with iterative attention}, booktitle={Int. Conf. Mach. Learn.}, pages={4651--4664}, year={2021}}

@article{han2026chemfusion, author={Han, Qiwei and others}, title={ChemFusion: A Multimodal Cross-Attention Network for Reaction Yield Prediction}, journal={arXiv preprint arXiv:2607.17033}, year={2026}}

@misc{rdkit, author={Landrum, Greg and others}, title={RDKit: Open-source cheminformatics software}, year={2024}}

@inproceedings{he2016deep, author={He, Kaiming and others}, title={Deep residual learning for image recognition}, booktitle={Proc. IEEE Conf. Comput. Vis. Pattern Recognit.}, pages={770--778}, year={2016}}

@inproceedings{wang2020what, author={Wang, Weiyao and others}, title={What makes training multi-modal classification networks hard?}, booktitle={Proc. IEEE/CVF Conf. Comput. Vis. Pattern Recognit.}, pages={12695--12705}, year={2020}}

@inproceedings{wu2022characterizing, author={Wu, Nan and others}, title={Characterizing and overcoming the greedy nature of learning in multi-modal representations}, booktitle={Int. Conf. Learn. Represent.}, year={2022}}

@article{fisher2019all, author={Fisher, Aaron and others}, title={All models are wrong, but many are useful: Learning a variable's importance}, journal={J. Mach. Learn. Res.}, volume={20}, pages={1--81}, year={2019}}

@article{ebejer2012freely, author={Ebejer, Jean-Paul and others}, title={Freely available conformer generation methods: how good are they?}, journal={J. Chem. Inf. Model.}, volume={52}, pages={1146--1158}, year={2012}}

@article{axelrod2022geom, author={Axelrod, Simon and others}, title={GEOM, energy-annotated molecular conformations}, journal={Sci. Data}, volume={9}, pages={185}, year={2022}}

@inproceedings{carion2020end, author={Carion, Nicolas and others}, title={End-to-end object detection with transformers}, booktitle={Eur. Conf. Comput. Vis.}, pages={213--229}, year={2020}}
		
	\end{multicols}
	
\end{document}